\documentclass[runningheads]{llncs}
\usepackage{graphicx}
\usepackage{comment}
\usepackage{amsmath,amssymb} % define this before the line numbering.
\usepackage{color}

\usepackage{helvet} % DO NOT CHANGE THIS
\usepackage{courier}  % DO NOT CHANGE THIS
\usepackage{graphicx}  % DO NOT CHANGE THIS
\usepackage{epsfig}
\usepackage{amsmath}
\usepackage{amssymb}
\usepackage{bm}
\usepackage{subfigure}
\usepackage{booktabs}
\usepackage{multirow}
\usepackage{booktabs}
\usepackage{caption}
\usepackage{float}
\usepackage{algorithm, algorithmic}
\usepackage{bbding}
\usepackage{array}
\usepackage{color}

\begin{document}
% \linenumbers
\pagestyle{headings}
\mainmatter
\def\ECCVSubNumber{1792}  % Insert your submission number here

\title{Wavelet-Based Dual-Branch Network \\for Image Demoir\'{e}ing} % Replace with your title

\begin{comment}
\titlerunning{ECCV-20 submission ID \ECCVSubNumber} 
\authorrunning{ECCV-20 submission ID \ECCVSubNumber} 
\author{Anonymous ECCV submission}
\institute{Paper ID \ECCVSubNumber}
\end{comment}
%******************

%\begin{comment}
\titlerunning{Wavelet-Based Dual-Branch Network for Image Demoir\'{e}ing}
\author{
Lin Liu\inst{1,2} \quad
Jianzhuang Liu\inst{2} \quad
Shanxin Yuan\inst{2}\thanks{Corresponding author} \quad 
Gregory Slabaugh\inst{2} \quad \\ 
Ale\v{s} Leonardis\inst{2} \quad
Wengang Zhou\inst{1} \quad
Qi Tian\inst{3}
}
\authorrunning{L. Liu et al.}
\institute{University of Science and Technology of China \\
%\url{http://www.springer.com/gp/computer-science/lncs} 
\and
Noah's Ark Lab, Huawei Technologies\\
\and
Huawei Cloud BU
}
%\end{comment}
%******************
\maketitle

\begin{abstract}

% problem
When smartphone cameras are used to take photos of digital screens, usually moir\'{e} patterns result, severely degrading photo quality.
% our paper want to do
In this paper, we design a wavelet-based dual-branch network (WDNet) with a spatial attention mechanism for image demoir\'{e}ing.
% why existing networks do not work well
Existing image restoration methods working in the RGB domain have difficulty in distinguishing moir\'{e} patterns from true scene texture.
% our proposal
Unlike these methods, our network removes moir\'{e} patterns in the wavelet domain to separate the frequencies of moir\'{e} patterns from the image content. The network combines dense convolution modules and dilated convolution modules supporting large receptive fields. 
% our method works
Extensive experiments demonstrate the effectiveness of our method, and we further show that WDNet generalizes to removing moir\'{e} artifacts on non-screen images. Although designed for image demoir\'{e}ing, WDNet has been applied to two other low-level vision tasks, outperforming state-of-the-art image deraining and deraindrop methods on the Rain100h and Raindrop800 data sets, respectively.

\end{abstract}

%%%%%%%%% BODY TEXT
\section{Introduction}

A smartphone has become an indispensable tool in daily life, and the popularity of mobile photography has grown supported by advancements in photo quality. It has become increasingly common to take photos of digital screens in order to quickly save information. 
However, when a photo is taken of a digital screen, moir\'{e} patterns often appear and contaminate the underlying clean image. These moir\'{e} patterns are caused by the interference between the camera's color filter array (CFA) and the screen's subpixel layout. 
In general, moir\'{e} patterns are likely to appear when two repetitive patterns interfere with each other. For example, moir\'{e} artifacts can appear when the repetitive patterns in textiles and building's bricks interfere with camera's CFA. 
Removing moir\'{e} patterns is challenging, as moir\'{e} patterns are irregular in shape and color, and can span a large range of frequencies. Unlike other image restoration tasks, such as image denoising \cite{zhang2017beyond,lefkimmiatis2018universal,tai2017memnet}, image demosaicing~\cite{gharbi2016deep,liu2020joint} and super resolution \cite{zhang2018residual,Jang_2019_CVPR_Workshops,huang2017wavelet,isobe2020video}, where the challenge is mainly in removing high-frequency noise and recovering details, image demoir\'{e}ing requires not only recovering high frequency image details, but also removing moir\'{e} patterns with frequencies spanning a large range.

\begin{figure}[!t]
		\centering
		\includegraphics[width=1.0\textwidth]{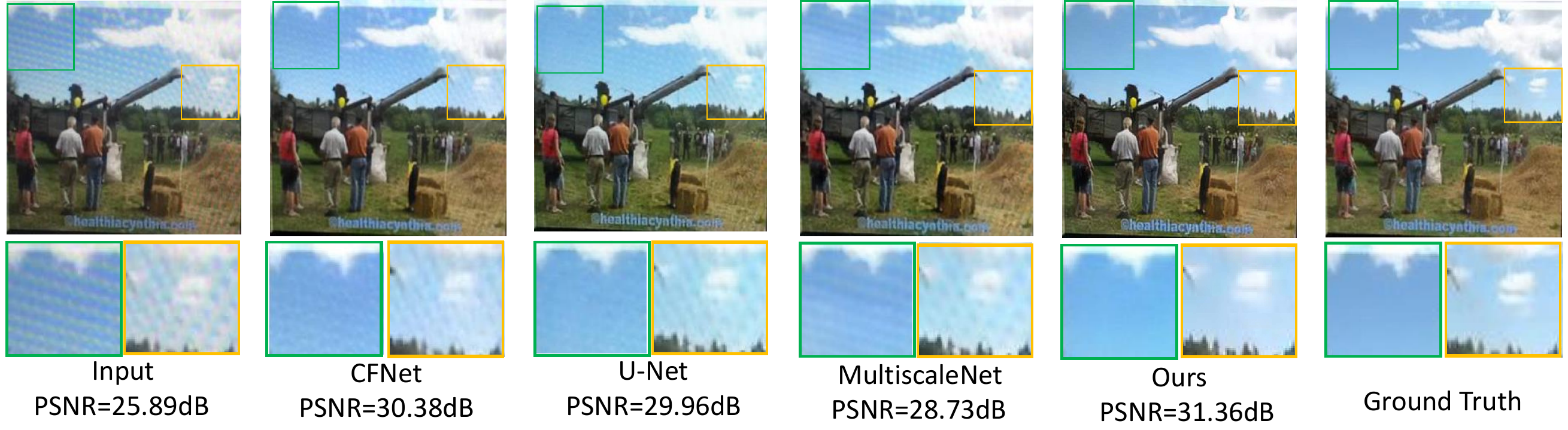}
		\caption{Qualitative comparison among CFNet~\cite{liu2018demoir}, U-Net~\cite{ronneberger2015u}, MultiscaleNet~\cite{sun2018moire} and our network on the TIP2018 data set~\cite{sun2018moire}. Ours is trained in the wavelet domain and removes the moir\'{e} patterns significantly better.}
		\label{index}
\end{figure}

Most existing image restoration methods \cite{ronneberger2015u,isola2017image,zhang2018residual} working in the RGB domain are not tailored to image demoir\'{e}ing, though a few attempts~\cite{sun2018moire,liu2018demoir,he2019mop} have been made to tackle it recently. 
Sun \textit{et al.}~\cite{sun2018moire} proposed a multi-resolution network to remove moir\'{e} patterns of different frequencies/scales. 
Liu \textit{et al.}~\cite{liu2018demoir} developed a coarse-to-fine network and trained it on a synthetic dataset and refined real images with GAN.
He \textit{et al.}~\cite{he2019mop}  utilized the edge information and appearance attributes of moir\'{e} patterns to demoir\'{e}.
However, all the existing methods (working in the RGB domain) have difficulty in distinguishing moir\'{e} patterns from the real image content, and in dealing with low-frequency moir\'{e} patterns~\cite{sun2018moire}. 
Fig.~\ref{index} shows some qualitative comparisons among four methods (including ours). 
We argue that image demoir\'{e}ing can be more easily handled in the frequency domain. Wavelet-based methods have been explored in computer vision and shown good performance, e.g., in classification \cite{fujieda2017wavelet,oyallon2017scaling}, network compression \cite{levinskis2013convolutional,gueguen2018faster}, and face super-resolution~\cite{huang2017wavelet}. However, due to their task-specific design, these methods cannot be directly used for image demoir\'{e}ing.

In this paper, we propose to remove moir\'{e} patterns in the frequency domain, where the input image with moir\'{e} patterns is first decomposed into different frequency bands using a wavelet transform.
After the wavelet transform, moir\'{e} patterns are more apparent in certain wavelet subbands, where they can be more easily removed (see Fig. \ref{fig:wavelet} for example). 
Our model, working in the frequency domain, is a dual-branch network with dense branches and dilated branches, which are responsible for restoring the close-range and far-range information, respectively.
We also design a spatial attention mechanism called a \emph{direction perception module} in the dense branches to highlight the areas with moir\'{e} patterns.

In this paper, we make following contributions:
\begin{enumerate}
\item We propose a novel wavelet-based and dual-branch neural network for image demoir\'{e}ing. We also propose a spatial attention mechanism called direction perception module (DPM) to highlight the areas with moir\'{e} patterns.
\item Our network achieves the best results on the demoir\'{e}ing task. Our trained model can also remove moir\'{e} artifacts on non-screen images.
\item Our new architecture generalises well to other low-level vision tasks, such as deraining and deraindrop, where we also obtain state-of the-art results.
\item In addition, we built a new  urban-scene data set with more types of moir\'{e} patterns, which will be made publicly available.
\end{enumerate}	
\section{Related Work}

In this section, we give a brief review of the most relevant work for image demoir\'{e}ing. 

\textbf{Moir\'{e} Pattern Removal.} 
% some old fashioned demoireing models
Early image demoir\'{e}ing work~\cite{sidorov2002suppression,sasada2003stationary,siddiqui2009hardware,liu2015moire,yang2017textured} focused on certain specific moir\'{e} patterns. 
Sidorov \textit{et al.}~\cite{sidorov2002suppression} presented a spectral model to address monotonous and monochrome moir\'{e} patterns. 
Other work tried to remove striped moir\'{e} artifacts \cite{sasada2003stationary,wei2012median,sidorov2002removing} or dotted artifacts \cite{siddiqui2009hardware,sun2014scanned,kim2018deep} in scanned or X-ray images.
Yang \textit{et al.}~\cite{yang2017textured} and Liu \textit{et al.}~\cite{liu2015moire} removed moir\'{e} artifacts using low-rank and sparse matrix decomposition. But their methods focus on textile moir\'{e} patterns and cannot handle low-frequency moir\'{e} patterns.
Compared to these methods, our approach can deal with a much larger range of moir\'{e} patterns as we do not make presumptions about the moir\'{e} patterns.
% some contemporary models
Contemporary models \cite{sun2018moire,liu2018demoir,he2019mop,zheng2020image} and recent challenges \cite{yuan2019aim,yuan2019aimresults,Yuan2020CVPRWorkshops} cast image demoir\'{e}ing as an image restoration problem addressed using deep learning. 
Recently Liu \textit{et al.}~\cite{liu2018demoir} built a coarse-to-fine convolutional neural network to remove moir\'{e} patterns from photos taken of screens. But the method mainly used synthetic data for network training and focuses on removing moir\'{e} patterns in text scenes. 
Sun \textit{et al.}~\cite{sun2018moire} proposed a multi-resolution convolutional neural network for demoir\'{e}ing and released an associated dataset. 
He \textit{et al.}~\cite{he2019mop} labeled the data set in \cite{sun2018moire} with three attribute labels of moir\'{e} patterns, which is beneficial to learn diverse patterns.
This method, along with \cite{liu2018demoir,sun2018moire}, works in the RGB domain and in a multi-scale manner. Such a strategy has limited capacity to correctly distinguish moir\'{e} patterns from true image content. In contrast, our method works in the wavelet domain, and we introduce a novel network design, resulting in stronger moir\'{e} pattern removal while restoring image details.

\textbf{Wavelet-based Methods.} 
Wavelet-based methods have been explored in some computer vision tasks, including classification \cite{fujieda2017wavelet,oyallon2017scaling,li2020wavelet,williams2018wavelet}, network compression \cite{levinskis2013convolutional,gueguen2018faster}, face aging \cite{liu2019attribute}, super-resolution \cite{huang2017wavelet,LiuMultiLevel}, style transfer \cite{yoo2019photorealistic}, etc. Fujieda \textit{et al.}~\cite{fujieda2017wavelet} proposed wavelet CNNs that utilize spectral information to classify the textures. Liu \textit{et al.}~\cite{liu2019attribute} used a wavelet-based method to capture age-related texture details at multiple scales in the frequency domain.
Our work is the first to use a wavelet transform to remove moir\'{e} patterns in the frequency domain.
Compared with Fourier transform and discrete cosine transform, wavelet transform considers both spatial domain information and frequency domain information.
With the wavelet composition, different wavelet bands represent such a broad range of frequencies, which can not be achieved by a few convolutional layers even with large kernels.

The most relevant in the above studies is \cite{huang2017wavelet} for using wavelets for face super-resolution, where a neural network is deployed to predict the wavelet coefficients. But predicting the wavelet coefficients of a moir\'{e}-free image from its moir\'{e} image in the RGB domain is difficult. Moir\'{e} patterns cover a wide range in both space and frequency domains, making it hard to distinguish moir\'{e} patterns from true scene textures.
% while our network runs in the wavelet domain. And 
For photo-realistic style transfer, Yoo \textit{et al.}~\cite{yoo2019photorealistic} regarded the wavelet transform and wavelet inverse transform as a pooling layer and an unpooling layer respectively% in the network
, with the aim of preserving their structural information. However, this technique~\cite{yoo2019photorealistic} may not be suitable for image demoir\'{e}ing, where moir\'{e} patterns can have strong structural information overlaid with true image contents.
Our work runs directly in the wavelet domain. Wavelet transform and inverse wavelet transform are used outside of the network. 

\textbf{Dual-branch Design.} 
Dual-branch network structure design makes use of two branches, complementing each other \cite{gao2017dual,qin2019difficulty,Fu_2019_CVPR,Si_2018_CVPR}.
It has been used in image super-resolution \cite{qin2019difficulty}, classification \cite{gao2017dual}, segmentation \cite{Fu_2019_CVPR} and person-re-ID \cite{Si_2018_CVPR}. 
Gao \textit{et al.}~\cite{gao2017dual} proposed a dual-branch network for polarimetric synthetic aperture radar image classification. One branch is used to extract the polarization features and the other to extract the spatial features.
Fu \textit{et al.}~\cite{Fu_2019_CVPR} proposed a dual attention network with a position attention module and a channel attention module for scene segmentation.
For image super-resolution, the dual branch in \cite{qin2019difficulty} can give consideration to easy image regions and hard image regions at the same time. 
All existing dual-branch networks have different branches, each focusing on certain image features, which only merge at the end of the network. In our design, the two branches iteratively communicate with each other to achieve a better representation. 

\textbf{Texture Removal.} 
Texture removal~\cite{subr2009edge,ham2015robust,zhang2014rolling} is related to demoir\'{e}ing, as moir\'{e} patterns can be viewed as a special type of texture.
Methods using local filters attempted to remove certain textures while maintaining other high-frequency structures \cite{subr2009edge,ham2015robust,zhang2014rolling}. 
Subr \textit{et al.}~\cite{subr2009edge} smoothed texture by averaging two manifolds from local minima and maxima. 
Both Ham \textit{et al.}~\cite{ham2015robust} and Zhang \textit{et al.}~\cite{zhang2014rolling} used dynamic guidance and a fidelity term for image filtering. Multiple strategies to smooth texture were used in \cite{su2012edge,karacan2013structure,bao2013tree}. 
Optimization-based methods were also exploited~\cite{sun2017image,xu2012structure,xu2011image}. 
Sun \textit{et al.}~\cite{sun2017image} and Xu \textit{et al.}~\cite{xu2011image} used $l_{0}$ minimization to retrieve structures from images. 
Xu \textit{et al.}~\cite{xu2012structure} optimized a global function with relative total variation regularization. 
However, these texture removal methods focus on the structure of the image and remove some repeated and identical or similar patterns. They do not address moir\'{e} patterns that span a much wider range of frequencies and appear in varied shapes and directions.

\section{Our Method}

\begin{figure*}[t]
  \centering
  \includegraphics[width=12cm]{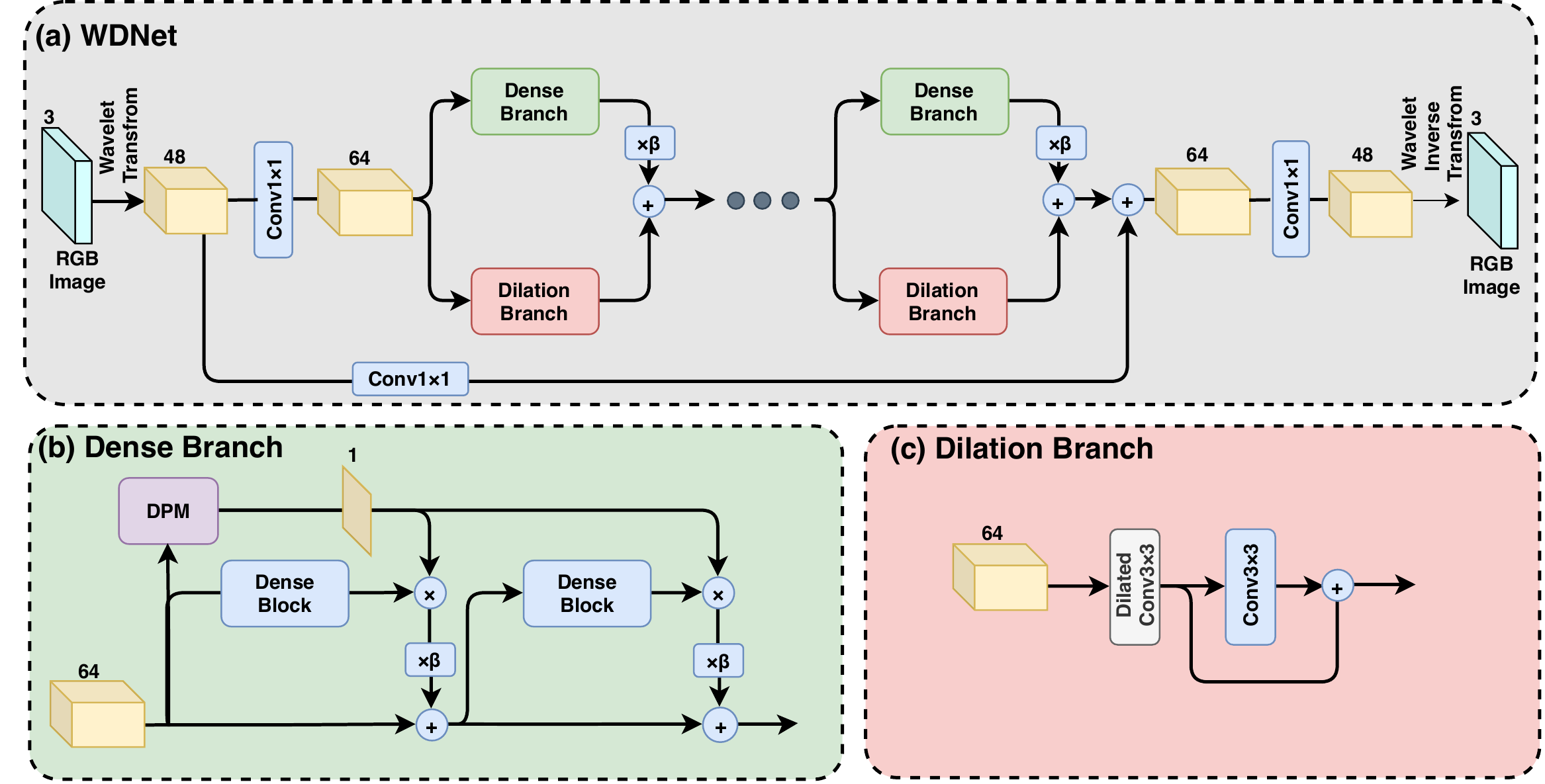}
  \caption{The architecture of the proposed WDNet (a) consisting of two kinds of branches, the dense branch (b) for restoring the close-range information and the dilation branch (c) for restoring the far-range information. Working in the wavelet domain can better remove the moir\'{e} patterns and retain the details. The numbers of channels are given in the figure.}
  \label{fig:network}
\end{figure*}

The architecture of \textbf{WDNet} is shown in Fig.~\ref{fig:network}. 
% general input, output
The original RGB image input is first transformed into the \textbf{W}avelet domain, where a \textbf{D}ual-branch \textbf{Net}work is used to remove the moir\'{e} patterns. The final RGB image output is obtained by applying the inverse wavelet transform to the network's output. 
% dual-branch
The dual-branch network has seven dual-branch modules, each including a dense branch and a dilation branch.
A ResNet~\cite{he2016deep} style skip connection is made. %
The dense and dilation branches are responsible for acquiring moir\'{e} information in different frequency ranges; the former detects close-range patterns and the latter detects far-range patterns.

\subsection{Working in the Wavelet Domain}

Our network operates in the wavelet domain to remove moir\'{e} patterns.
We employ 2D fast wavelet transform (FWT) to decompose the input RGB image into a sequence of wavelet subbands (coefficients) of a common smaller size, but with different frequency content. 
We choose the simplest Haar wavelet as the basis for the wavelet transform.  A Haar wavelet transform can be efficiently computed and suitable for our task.
The FWT iteratively applies low-pass and high-pass decomposition filters along with downsampling to compute the wavelet coefficients where the low-pass filter = $\left(1/\sqrt{2}, 1/\sqrt{2}\right)$ and the high-pass filter = $\left(1/\sqrt{2},-1/\sqrt{2}\right)$. 
In each level of the transform, we use the high-pass filter and the low-pass filter along the rows to transform an image to two images, and then we apply the same filters along the columns of these two images, obtaining four images in total.
The equations to derive the subbands can be found in \cite{Rafael}.

\begin{figure}[t]
  \centering
  \includegraphics[width=1.0\textwidth]{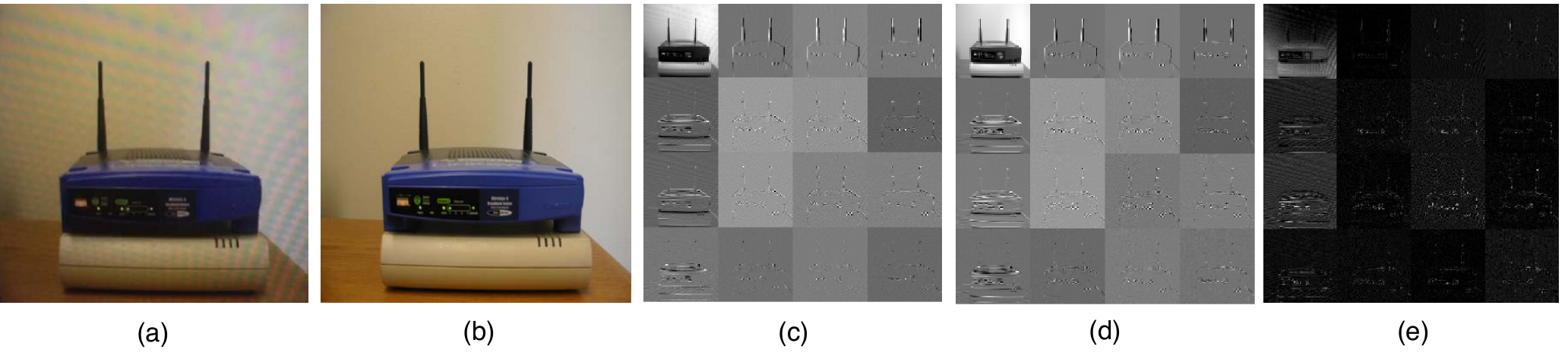}
  \caption{(a) An image with moir\'{e} patterns. (b) The ground truth of (a). (c) Wavelet subbands transformed from the gray-level image of (a). (d) Wavelet subbands transformed from the gray-level image of (b). (e) Difference between (c) and (d).}
  \label{fig:wavelet}
\end{figure}

As shown in Figs. \ref{fig:wavelet}(c) and (d), the wavelet subbands at different levels correspond to different frequencies. 
Moir\'{e} patterns are conspicuous only in certain wavelet subbands.
For example, in Fig.~\ref{fig:wavelet}(e), the difference between (c) and (d) shows that the first three wavelet subbands of the first column in Fig.~\ref{fig:wavelet}(c) contain obvious and different moir\'{e} patterns. 
From our experiments, we find that wavelet subbands with higher frequencies usually contain fewer moir\'{e} patterns. These wavelet subbands provide important information to recover the details for moir\'{e} pattern-free images.

In our method, the wavelet transform level is set to 2.
The original RGB input is transformed into 48 subbands, with 16 subbands for each color channel.
As such, another benefit of working in the wavelet domain is that the spatial size of the original image $\left(H\times W \times 3\right)$ is reduced by a quarter in both the width and height $((H/4) \times (W/4) \times 48)$.  The reduced spatial size consequently reduces the computation in the deep network.
Besides, we concatenate all the 48 channels instead of processing the high and low frequency bands individually, because moire patterns for each image vary a lot, it is difficult to find a threshold to manually separate them.

\subsection{Dense Branch}
DenseNet~\cite{huang2017densely} can alleviate the problem of vanishing gradient and reduce the number of network parameters through the design of bypass connections and feature reuse. 
In low-level vision, dense blocks have been used in deraining and dehazing \cite{shen2018deep,guo2019dense123}, and image super-resolution with the Residual Dense Network (RDN)~\cite{zhang2018residual} or other networks \cite{Jang_2019_CVPR_Workshops}. 

As shown in Fig.~\ref{fig:network}(b), we design each dense branch by adding a new Direction Perception Module (DPM) to the residual dense module borrowed from RDN~\cite{zhang2018residual}. 
The residual dense module has two small dense blocks each with 5 convolutional layers. In the same dense block, the input of each layer contains the output of all previous layers. DPM is used to find the directions of moir\'{e} patterns and will be explained in Section~\ref{sec:dpm} in detail. The output of DPM and each dense block are multiplied, weighted by a factor $\beta$, and then the result is added to the input. This design can effectively find the locations of close-range moir\'{e} patterns.

\subsection{Dilation Branch}

Subsampling or pooling feature maps can enlarge receptive fields, but loses some details. %
Dilated convolution is used to overcome this problem. 
In each of our dilation branches (Fig. \ref{fig:network}(c)), there are two layers: a $3\times3$ dilated convolution layer with a dilation rate $d_{k+2}=d_{k}+d_{k+1}$, followed by a traditional $3\times3$ convolution, where $k$ is the index of a dilation branch with $d_{1}=1$ and $d_{2}=2$. The rationale for this design is explained below. 

When multiple dilated convolutions with a fixed dilation rate (say, 2) are applied successively, many pixels are not involved in the convolutions and a special kind of artifact called \emph{gridding} appears \cite{wang2018understanding}. 
Alternatively, a hybrid design using dilation rates (1,2,3,1,2,3,...)~\cite{wang2018understanding} is not suitable for our task because these dilations are not large enough to cover general moir\'{e} patterns. Therefore, we design our dilation rates according to the Fibonacci series (e.g., (1,2,3,5,8,13,21) when there are 7 dilation branches). Additionally, we apply a traditional $3\times3$ convolution on the output of the dilated convolution to further reduce any gridding artifact (Fig. \ref{fig:network}(c)).

\begin{figure}[tb]
  \centering 
  \includegraphics[width=12.2cm]{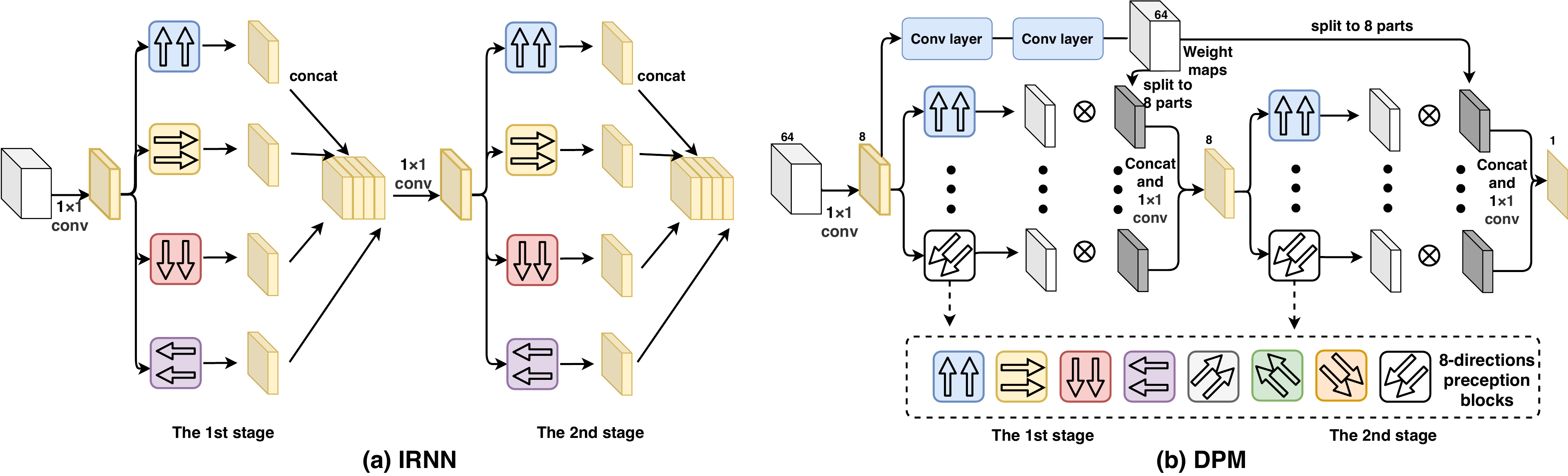}
  \caption{(a) Image recurrent neural network (IRNN). (b) Our direction perception module (DPM) with two stages, each with 8 directions.}
  \label{fig:dpm}
\end{figure}

\subsection{Direction Perception Module}
\label{sec:dpm}
We propose a Direction Perception Module (DPM) by improving the Image Recurrent Neural Network (IRNN) \cite{bell2016inside,Xu_2018_ECCV,hu2019direction}, as shown in Fig. \ref{fig:dpm}.
The two-stage four-directional
IRNN (Fig. \ref{fig:dpm}(a)) architecture enhances the use of contextual information, where the first stage in IRNN aims to produce feature maps that acquire neighboring contextual information and the second stage in IRNN further gathers the global information. 
However, in IRNN, in the feature maps at the end of the first stage, a pixel can only obtain information in the vertical and horizontal directions, but not in other directions. %

As shown in Fig. \ref{fig:dpm}(b), we improve IRNN in two aspects.
First, we extend the 4-direction perception to 8-direction perception by including 4 more diagonal directions
to enhance the detection ability of slanting moir\'{e} patterns.
The moir\'{e} patterns in other directions are detected in the second stage. 
Second, the weight maps are used to distinguish the importance of information in different directions.
Unlike IRNN which learns direction-aware features in the embedded space, we use DPM to generate the attention map which highlights moir\'{e} pattern spatial distributions and is supervised by $L_{a}$ discussed in Sec. 3.6.
The combination of the two convolution layers and the eight-direction perception blocks is equivalent to an attention operation.

\subsubsection{Loss Function.} 
%In this section, we describe the loss function. 
The loss function of the whole network is defined as follows:
\begin{equation}
L = \lambda_{1} L_{a}+\lambda_{2} L_{l _{1}}+\lambda_{3} L_{p}+\lambda_{4} L_{w}{,}
\label{fml:lossf}
\end{equation}	
where $L_{l_{1}}$ is the $l_{1}$ loss between the image and its ground-truth in the RGB domain,
$L_{p}$ is the perceptual loss \cite{johnson2016perceptual}, %
and the other parts are described below.

\subsubsection{Wavelet Loss.}
Let $C=(c_{1},c_{2},...,c_{N})$ and $\hat{C}=(\hat{c}_{1},\hat{c}_{2},...,\hat{c}_{N})$ be the ground-truth and the estimated wavelet subbands, respectively. 
Two wavelet-based losses, MSE loss $l_{\text {MSE}}$ and detail loss $l_{detail}$, are defined as follows:
\begin{equation}
l_{\text {MSE}}=\gamma_{1}\sum_{i=1}^{3}\left\|\hat{c}_{i}-c_{i}\right\|^{2}+\gamma_{2}\sum_{i=4}^{N} \left\|\hat{c}_{i}-c_{i}\right\|^{2}{,}
\label{eq:wavemseloss}
\end{equation}
where the first term is for the low-frequency components, and the second for the high-frequency ones, and $\gamma_{1}$ and $\gamma_{2}$ are two weighting factors.
\begin{equation}
l_{detail}=\sum_{i=4}^{N} \max \left(\alpha \left|c_{i}\right|^{2}-\left|\hat{c}_{i}\right|^{2}, 0\right){,}
\end{equation}
which is inspired by the work in super-resolution \cite{huang2017wavelet} to prevent the wavelet coefficients from converging to 0 and therefore retaining the high-frequency details of the image. $\alpha$ is a positive number and set to 1.2 in this work. 
Setting $\alpha$ to a value greater than 1 can effectively prevent $\hat{c}_{i}$ from converging to 0.
Different from the loss function in \cite{huang2017wavelet}, ours does not have a bias value $\epsilon$, because we found that $\epsilon$ has little effect on the final results. Besides, the value of $\epsilon$ in \cite{huang2017wavelet} was set to 0 in their source code.
The network's wavelet loss $L_{w}$ is then:
\begin{equation}
L_{w}=l_{\text {MSE}}+ l_{detail}{.}
\label{eq:waveloss}
\end{equation}

\subsubsection{Attention Loss.}
We further apply the attention loss $L_{a}$ as:
\begin{equation}
L_{a }=\|A-d\left(M\right)\|_{2}^{2},
\end{equation}
where $A$ is the output of the DPM in the network and $M$ is the mask of the moir\'{e} patterns, which is computed by thresholding the difference between the moir\'{e}-contaminated image and the ground-truth. $d(\cdot)$ is a downsampling operation to ensure that the mask and $A$ have the same size. The threshold is set to 15 in this work.

\section{Experiments}

We compare WDNet with state-of-the-art methods, and conduct user studies for image demoir\'{e}ing. We also demonstrate that WDNet when trained only for screen image demoir\'{e}ing can also remove moir\'{e} artifacts on non-screen images.
Finally, we apply WDNet to two other tasks, deraining and de-raindrop, to show that our method generalizes well to other frequency-based low-level vision tasks.

\subsection{Implementation Details}

Throughout our experiments, the input is a $256\times256\times3$ RGB image, which is then transformed into 48 different-level wavelet subbands with the size of $64\times64$ by a Haar wavelet transform. 
We use seven dual-branch modules and add DPM into the 4th dense branch to perceive moir\'{e} patterns.
Our model is implemented in PyTorch and runs on a NVIDIA Tesla V100 GPU.
The batch size is set to 16 during training. Adam \cite{kingma2014adam} is used to search the minimum of the loss function. The learning rate of the generator is 0.0002 and reduced by one-tenth for every 20 epochs.
In Eqns. (\ref{fml:lossf}) and (\ref{eq:wavemseloss}), the parameters $\lambda_{1}$, $\lambda_{2}$, $\lambda_{3}$, $\lambda_{4}$, $\gamma_{1}$ and $\gamma_{2}$ are empirically set to 1, 5, 10, 200, 0.01 and 1.1 respectively.
The method of \textit{eye} initialization in~\cite{zhang2018residual} is used to initialize the weights of WDNet. The value of $\beta$ throughout the paper is set to 0.2. 
The dense branch has more parameters than the dilation branch.
Setting $\beta$ to 0.2 makes the weights of the dilation branch’s parameters larger, which can maintain the details of the original image.

\subsection{Datasets and State-of-the-Art Methods} 

We perform experiments on two datasets, TIP2018~\cite{sun2018moire} and a new London's Buildings. The TIP2018 dataset contains 100,000+ image pairs with about $800\times800$ resolution and the moir\'{e}-free images come from the ImageNet ISVRC 2012 dataset.
The new dataset London's Buildings was built by us recently and will be publicly released. It is an urban-scene data set and its images contain bricks, windows and other regular patterns which are prone to generate moir\'{e} patterns. We use mobile phones and screens different from those used in the TIP2018 data set and thus provide additional diversity of moir\'{e} patterns. London's Buildings includes 400 training pairs and 60 testing pairs with about $2300\times1700$ resolution.
The $256\times256$ images used in training are resized (in TIP2018) or cropped (in London's buildings) from the original higher-resolution images.
We conduct two experiments on the TIP2018 data set, TestA and TestB. TestB follows the same setting as in \cite{sun2018moire}. TestA is a subset of TestB and this subset contains 130 hard samples from the same camera. 

We compare our model with MuiltiscaleNet~\cite{sun2018moire}, CFNet~\cite{liu2018demoir}, Pix2pix~\cite{isola2017image}, U-Net~\cite{ronneberger2015u}, MopNet~\cite{he2019mop}, and ResNet-34~\cite{he2016deep}. For fair comparison, when a network is trained on the TIP2018 data set, we use the same setting as in \cite{sun2018moire}. 
We faithfully reimplement MuiltiscaleNet~\cite{sun2018moire} and
CFNet~\cite{liu2018demoir} based on their papers.   
Note that since MopNet is very recent work without its code released, we can only compare with it in one experiment where \cite{he2019mop} gives its result on TIP2018.

\subsection{Comparison with the State-of-the-Art}
\begin{figure}[t]
  \centering
  \includegraphics[width=1.0\textwidth]{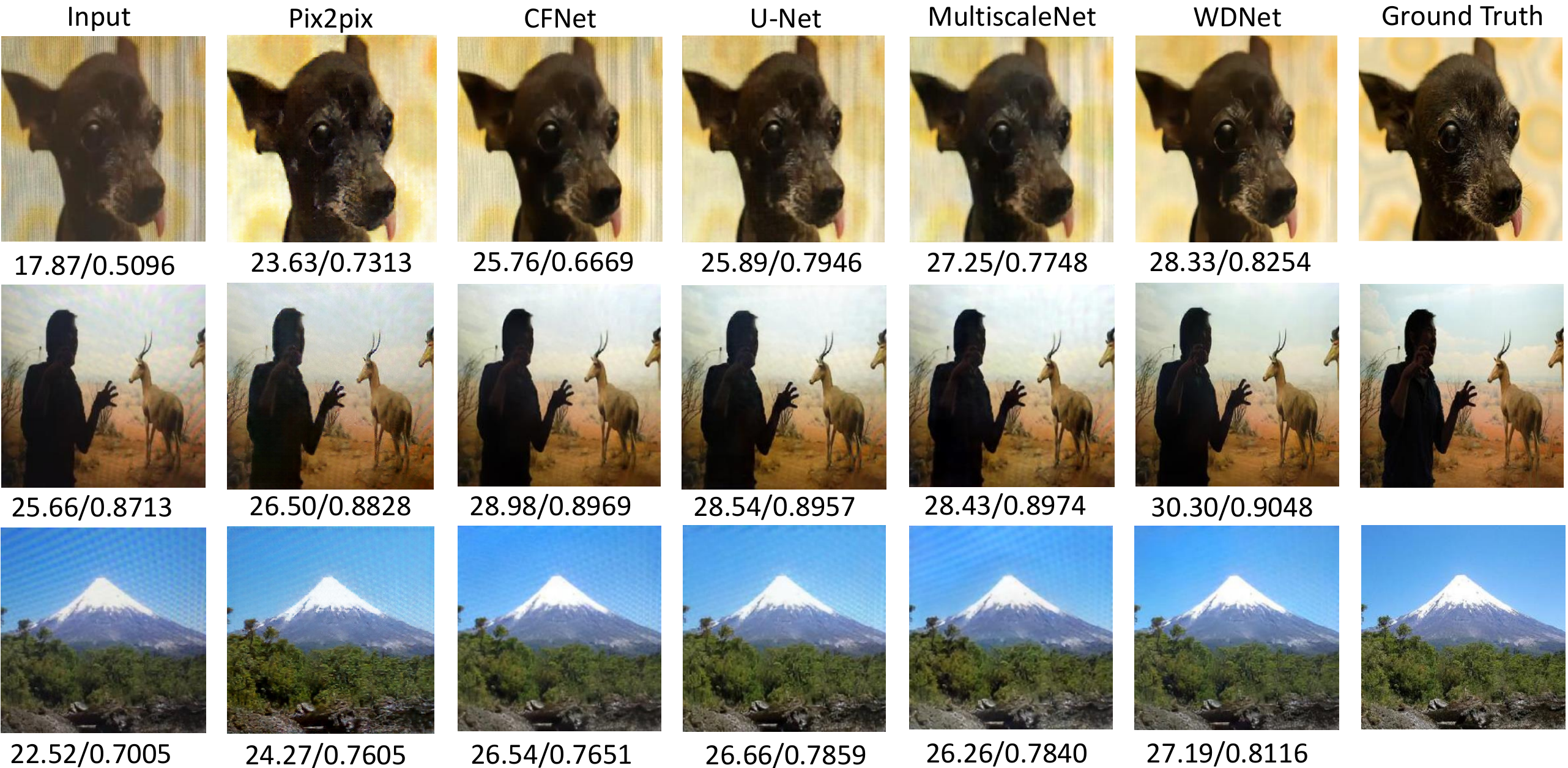}
  \caption{Visual comparison among Pix2pix~\cite{isola2017image}, CFNet~\cite{liu2018demoir}, U-Net~\cite{ronneberger2015u}, MuiltiscaleNet~\cite{sun2018moire} and our model WDNet, evaluated on images from the TIP2018 data set. The numbers under the images are PSNR/SSIM. }
  
  \label{fig:tip}

\end{figure}

\begin{figure}[t]
  \centering
  \includegraphics[width=1.0\textwidth]{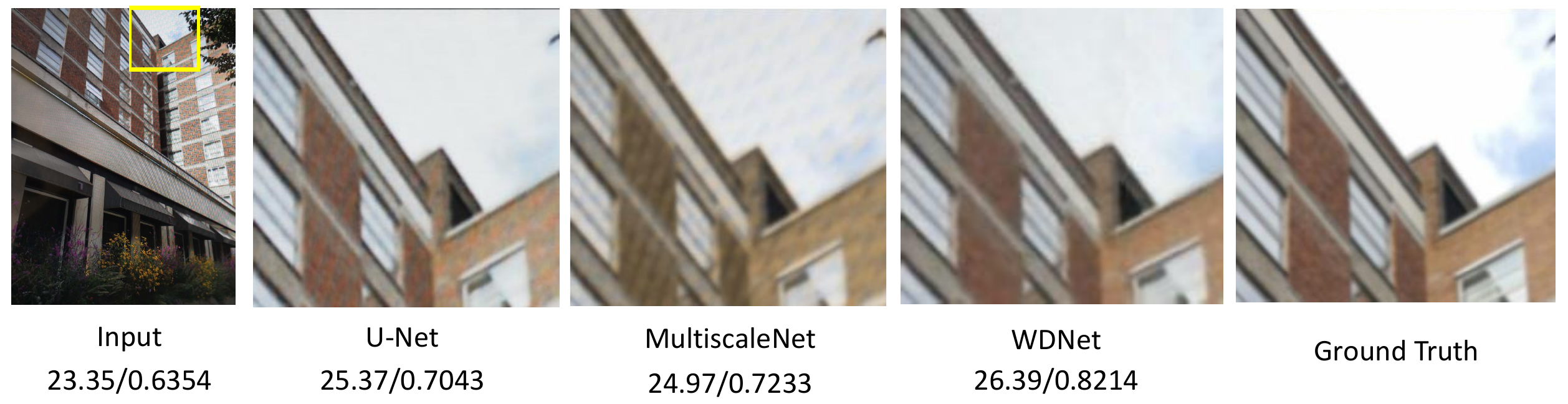}
  \caption{Visual comparison on an image from the  London's Building data set.}
  \label{fig:building}
\end{figure}

For quantitative comparison, we choose the PSNR and SSIM.
As shown in Tables \ref{tab:tip2018} and \ref{tab:london}, our method outperforms the state-of-the-art on both datasets. Note that our model outperforms the most recent MopNet \cite{he2019mop} by 0.33dB, even though MopNet uses additional annotation (labelling TIP2018 with three  attribute labels of moir\'{e} patterns) for training.
We show some qualitative results in Figs. \ref{fig:tip} and \ref{fig:building}, where there are different kinds of moir\'{e} patterns appearing in the input images. Our WDNet most effectively removes the moir\'{e} patterns.
For the computation consumption, CFNet, MultiscaleNet, U-Net, MopNet and WDNet take 58.2ms, 5.7ms, 3.8ms, 71.8ms and 14.6ms, respectively, to restore a $256\times256$ image on one NVIDIA Tesla P100 GPU.

\setlength{\tabcolsep}{4pt}
\begin{table}[t]
  \begin{center}
  \caption{Quantitative comparison on TIP2018. The best results are in \textbf{bold}. }
  \label{tab:tip2018}
  \resizebox{12.0cm}{!}{
  \begin{tabular}{ccccccccc }
    \toprule
      Test set &  & Pix2pix &U-Net&CFNet& MultiscaleNet& MopNet& WDNet\\
    \midrule 
    TestA&PSNR/SSIM&25.32/0.756&25.80/0.803&25.52/0.810&26.11/0.801&--/--& $\mathbf{27.88}$/$\mathbf{0.863}$\\
    \midrule 
    TestB&PSNR/SSIM&25.74/0.825&26.49/0.864&26.09/0.863&26.99/0.871&27.75/0.895&$\mathbf{28.08}$/$\mathbf{0.904}$\\
    
    \bottomrule
  \end{tabular}}
  \end{center}
\end{table}

\begin{table}[t]
  
  \centering\small
  \caption{Quantitative comparison on London's Building.}
  \label{tab:london}
  \resizebox{9.5cm}{!}{
  \begin{tabular}{cccccc }
    \toprule
    \ &	U-Net&	MultiscaleNet&	ResNet &CFNet&	WDNet \\                                       
    \midrule 
     
     PSNR/SSIM&23.48/0.790&	23.64/0.791&	23.87/0.780& 23.22/0.764	& $\mathbf{25.41} /\mathbf{0.839} $ \\
    \bottomrule
  \end{tabular}}
\end{table}

\begin{table}[t!]
  \centering\small
  \caption{User study. The preference rate shows the
percentage of comparisons in which the users prefer our results.}
  \label{tab:userstudy}
  \resizebox{8.5cm}{!}{
  \begin{tabular}{cccc}
    \toprule
      
      & WDNet$>$CFNet& WDNet$>$U-Net& WDNet$>$MultiscaleNet\\                                       
    \midrule 
   Preference rate & 83.7\% &  82.1\% &  80.5\%\\
    \bottomrule
  \end{tabular}}
\end{table}

\begin{figure}[t!]
  \centering
  \includegraphics[width=0.99\textwidth]{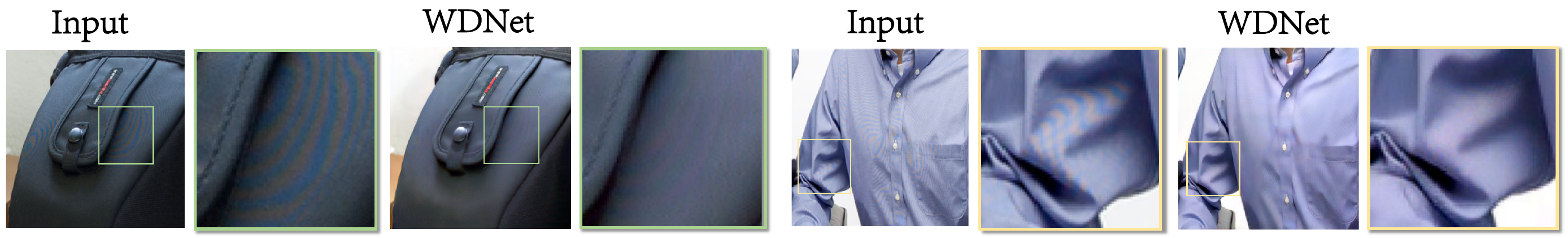}
  \caption{Some results on non-screen images.}
  \label{fig:nature}

\end{figure}

\begin{figure}[t]
  \centering
  \includegraphics[width=0.99\textwidth]{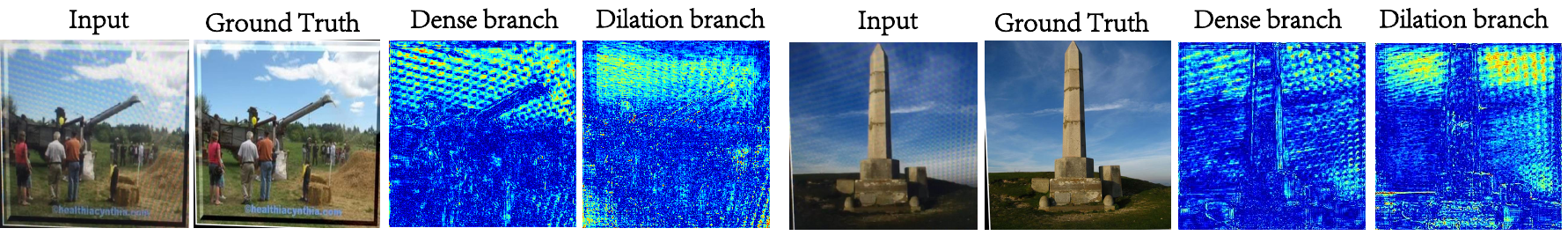}
  \caption{The heat maps of the feature maps outputted from the two branches of the 6th layer.}
  \label{fig:visualization map}
\end{figure}

\begin{figure}[t]
  \centering
  \includegraphics[width=1.0\textwidth]{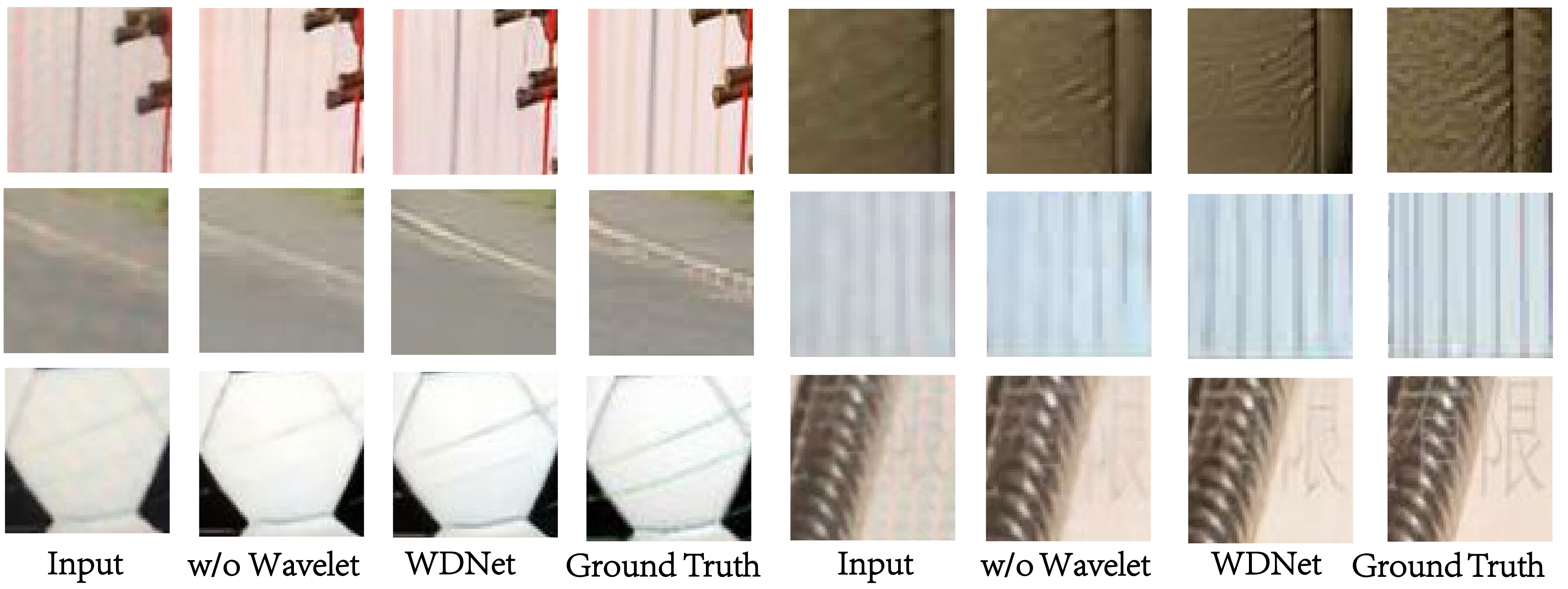}
  \caption{Visual comparison between WDNet and WDNet without wavelet transform. These patches show that the former restores the details better and removes more moir\'{e} patterns.}
  \label{fig:wrgb}
  
\end{figure}

\subsubsection{User Studies} 
We also conduct a user study with 10 users to compare the visual quality of the results generated by the models. During the study, two images $I_{1}$ and $I_{2}$, respectively obtained by our model and another model, and the corresponding moir\'{e}-free image (ground truth) $I$, are shown to a user who is asked to choose one from $I_{1}$ and $I_{2}$ which is closer to $I$.
There are 100 triplets of these images randomly selected from TIP2018 to compare two models and the results are reported in Table \ref{tab:userstudy}.

\subsubsection{Removal of Moir\'{e} Artifacts on Non-screen Images}
Moir\'{e} artifacts may also appear on objects with dense repetitive textures, such as buildings and clothes.
This kind of moir\'{e} artifacts usually does not always appear on the whole image.
In Fig. \ref{fig:nature}, we show some results of directly applying our WDNet model trained on TIP2018 without fine-tuning on non-screen images. It shows that our method is effective in removing this kind of moir\'{e} patterns as well.

\subsection{Ablation Studies}
To conduct ablation studies to quantify the contributions of the components of WDNet, the TestA dataset is used in the experiments.
To understand the restoration strategy between the two branches in the dual-branch architecture, we examine the feature maps of WDNet. At layer six, we normalize each channel of the feature map, and then sum up the values channel-wise to obtain a heat map. 
As shown in Fig. \ref{fig:visualization map}, the dense branch focuses on the local moir\'{e} patterns (interlaced bright and dark areas), while the dilation branch focuses on the whole areas contaminated by the moir\'{e} patterns due to its larger receptive field.
It demonstrates that the two types of branches are responsible for the acquisition of the close-range and far-range information and jointly restore the image.

To test the effectiveness of the dense branches, we replace the dense blocks of them with ResNet blocks.
To test the dilation branches, we replace the dilated convolutions with normal convolutions. 
As shown in Table \ref{tab:abstudy2}, both the dense and dilation branches play an important role in WDNet.
When the dense branches or the dilation branches are replaced, the PSNR decreases by 0.52dB or 1.05dB, respectively.
The PSNR decreases by 0.41dB when we remove the DPM module and the attention loss, and by 0.27dB when we remove the weight maps in DPM.
In addition, to compare the 8-directional DPM with the 4-directional IRNN, using the latter makes the PSNR decrease by 0.19dB. 

In order to verify whether working in the wavelet domain is better than
in the RGB domain, we design a network by removing the wavelet transform and wavelet inverse transform from WDNet, without the wavelet loss during training. These two networks are compared at the same computation amount and the results are shown in Table \ref{tab:abstudy2}. A visual comparison is shown in Fig. \ref{fig:wrgb}. From Table \ref{tab:abstudy2} and Fig. \ref{fig:wrgb}, we can see that the complete model best achieves moir\'{e} pattern removal. The wavelet transform can help to keep the edges and prevent the details from blurring.

The above studies clearly demonstrate that all the WDNet's components and the network design have their contributions to the excellent performance of WDNet. We also perform an experiment to explore the effect of different wavelet transform levels on the network's performance with the same network parameters. For levels 1, 2 and 3, the PSNR/SSIM are 27.24/0.866, 27.88/0.863, and 27.94/0.843, respectively. As the number of levels increases, the PSNR increases, but the SSIM decreases. The reason for such an interesting phenomenon will be left as the future work. 
Some recent related work such as \cite{blau2018perception} shows that perceptual quality and distortion (PSNR) may conflict with each other, particularly for image restoration tasks. %So for

\begin{figure}[t]
  \centering
  \includegraphics[width=1.0\textwidth]{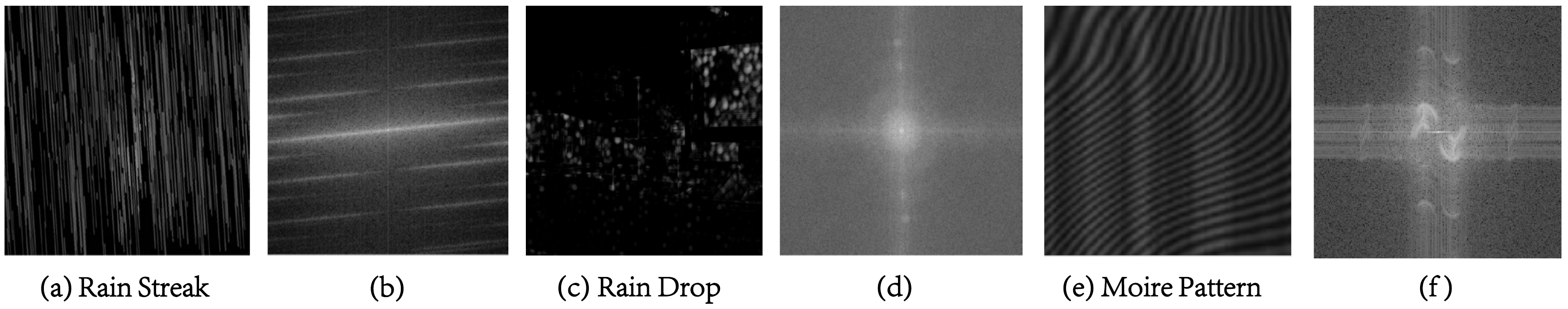}
  \caption{Frequency spectra of images of rain streak, rain drop and moir\'{e} pattern.}
  \label{fig:ft}
\end{figure}

\begin{table}[t]
  \centering\small
  \caption{Ablation study on TestA. (a) Replace dilated convolutions with normal convolutions. (b) Replace the dense blocks with the residual blocks. (c) Without the
  wavelet and inverse wavelet transfrom. (d) Without DPM. (e) Replace the 8-directional DPM with the  4-directional IRNN.}
  \label{tab:abstudy2}
  \resizebox{12.0cm}{!} {
  \begin{tabular}{lcccccc}
    \toprule
      Network &a &b&c&d&e&Complete model \\                                       
     
     \midrule 
     PSNR/SSIM &26.83/0.858&27.36/0.856&27.03/0.859&27.47/0.853&27.69/0.857&27.88/0.863\\

    \bottomrule
  \end{tabular}}

\end{table}

\subsection{Extension to Deraining and Deraindrop}

We show that our network also has advantages over the state-of-the-art methods in some other low-level vision tasks, such as deraining and deraindrop.
Deraining methods can be divided into traditional methods and deep-learning methods. Traditional methods \cite{luo2015removing,chang2017transformed} remove rain streaks in images by designing some hand-crafted priors. Deep-learning methods \cite{yang2017deep,li2018recurrent,Hu2019CVPR} use convolutional neural networks to learn the rain streak features automatically. Yang~\textit{et al.}~\cite{yang2017deep} created a multi-task network to jointly detect and remove rain. And Li~\textit{et al.}~\cite{li2018recurrent} proposed a recurrent neural network, where useful information in previous stages is beneficial to the rain removal in later stages. However, few deep-learning methods study how to use the feature of the rain streaks in the frequency domain to derain. In Fig. \ref{fig:ft}, similar to moir\'{e} patterns, rain streaks include high-frequency and low-frequency patterns, depending on the severity of the rain and distance to the camera. However, unlike moir\'{e} patterns, which can have different orientations in the same image, the directions of rain streaks are more consistent, resulting in an easier image restoration task. Like demoir\'{e}ing, after the wavelet transform, different types of rain steaks in the RGB domain are easier to distinguish in the wavelet domain.

\begin{figure}[t!]
  \centering
  \includegraphics[width=0.95\textwidth]{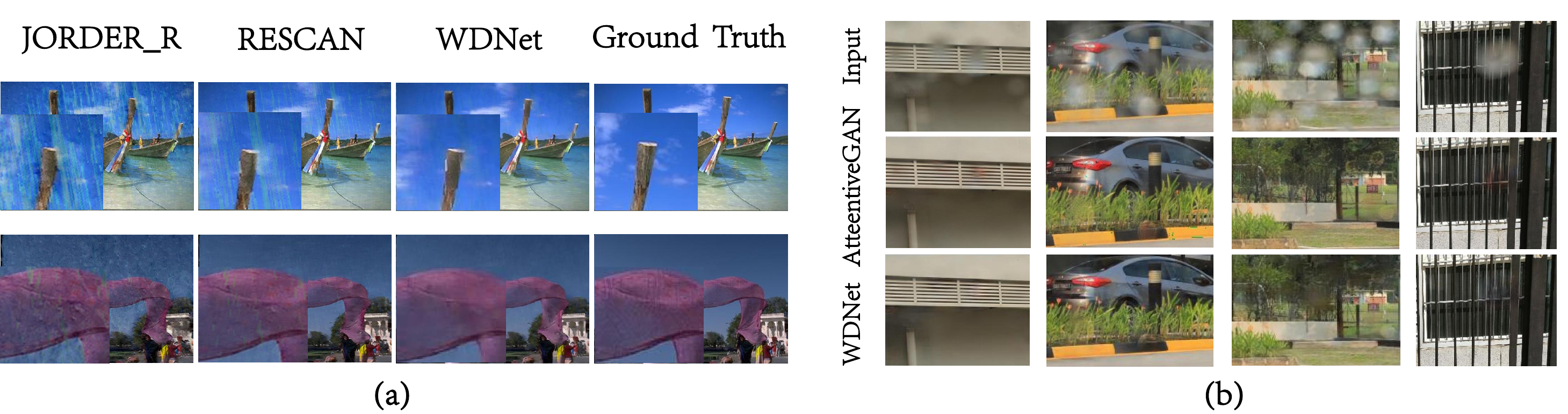}
  \caption{(a) Visual deraining comparison among JORDER\_R, RESCAN and WDNet. (b) Visual deraindrop comparison between AttentiveGAN and WDNet.}
  \label{fig:rain}
\end{figure}

We train WDNet on the Rain100h data set~\cite{yang2017deep} for deraining and on the Raindrop800 data set~\cite{qian2018attentive} for deraindrop, and test it on the corresponding test sets.
Though the textures of the rain/raindrop and moiré patterns are very different in color and structure, our WDNet method can also remove them well.
The results are given in Table \ref{tab:rain100h}, Table \ref{tab:raindrop} and Fig. \ref{fig:rain}, which show that our WDNet performs better than the state-of-the-art.
More visual results are given in the supplementary material.

\begin{table}[!t]
  \centering\small
  \caption{Quantitative comparison on Rain100h.}
  \label{tab:rain100h}
  \resizebox{12.0cm}{!}{
  \begin{tabular}{lcccccc}
    \toprule
    Method	& DSC~\cite{luo2015removing} &	JORDER\_R~\cite{yang2017deep}& DID\_MDN~\cite{zhang2018residual}&RESCAN~\cite{li2018recurrent}&DAFNet~\cite{Hu2019CVPR}&WDNet(Ours)\\
   \midrule
PSNR/SSIM &15.6/0.5446&23.45/0.749&24.53/0.799&26.45/0.845&28.44/0.874&$\mathbf{28.60}$/$\mathbf{0.878}$\\

    \bottomrule
  \end{tabular}}
\end{table}

\begin{table}[!t]
  \centering\small
  \caption{Quantitative comparison on Raindrop800.}
  
  \label{tab:raindrop}
  \resizebox{12.0cm}{!} {
  \begin{tabular}{lcccccc}
    \toprule
   	 Method &Eigen~\cite{eigen2013restoring}&Pix2pix~\cite{isola2017image}&$\mathrm{A^{2}Net}$~\cite{lin20182net}&AttentiveGAN~\cite{qian2018attentive}&Quan's~\cite{quan2019deep} &WDNet\ (Ours)\\
   	\midrule
   	PSNR/SSIM&23.90/0.80&28.85/0.84&30.79/0.93&31.52/0.92&31.44/$\mathbf{0.93}$&$\mathbf{31.75}$/$\mathbf{0.93}$\\

    \bottomrule
  \end{tabular}}
 
\end{table}

\section{Conclusion}
	
We have proposed a novel wavelet-based dual-branch network (WDNet) to remove moir\'{e} patterns. Working in the wavelet domain can restore more details and is more effective at moir\'{e} pattern removal than in the RGB domain. WDNet's dense branches and dilation branches are responsible for the acquisition of close-range and far range information, respectively. The new DPM module inside some dense branches can better perceive slanting moir\'{e} patterns. WDNet substantially outperforms other state-of-the-art models. Moreover, it obtains the best results in two other low-level vision tasks, deraining and deraindrop. In addition, we built a new urban-scene data set with challenging types of moir\'{e} patterns, which will be made publicly available. Our future work includes applications of WDNet on other vision tasks such as denoising and demosaicing.

\bibliographystyle{splncs04}
\bibliography{egbib}
\end{document}